\documentclass[10pt,twocolumn,letterpaper]{article}

\usepackage[final]{cvpr}      

\usepackage{url}
\usepackage{times}
\usepackage{soul}
\usepackage{url}
\usepackage[hidelinks]{hyperref}
\usepackage[utf8]{inputenc}
\usepackage{graphicx}
\usepackage{amsmath}
\usepackage{amsthm}
\usepackage{booktabs}
\usepackage{algorithm}
\usepackage{algorithmic}
\usepackage{verbatim}
\usepackage{multirow}
\usepackage{makecell}
\usepackage{enumitem}

\usepackage{tabularx}
\usepackage{ctable}
\usepackage{amssymb}
\usepackage[utf8]{inputenc} 
\usepackage[T1]{fontenc}    
\usepackage{hyperref}       
\usepackage{url}            
\usepackage{booktabs}       
\usepackage{amsfonts}       
\usepackage{nicefrac}       
\usepackage{microtype}      
\usepackage{xcolor}         
\usepackage{colortbl}
\usepackage{bbding}

\usepackage{pifont}
\usepackage{bbm}
\usepackage{url}
\usepackage{nicefrac}
\newcommand{\xmark}{\text{\ding{55}}}

\usepackage[marginal]{footmisc}
\newcommand{\typ}[1]{\textcolor{black}{#1}}
\newcommand{\zbs}[1]{\textcolor{black}{#1}}
\newcommand{\lyx}[1]{\textcolor{black}{#1}}
\newcommand{\leone}[1]{\textcolor{black}{#1}}
\newcommand{\zbsN}[1]{\textcolor{black}{#1}}
\usepackage[capitalize]{cleveref}
\crefname{section}{Sec.}{Secs.}
\Crefname{section}{Section}{Sections}
\Crefname{table}{Table}{Tables}
\crefname{table}{Tab.}{Tabs.}

\definecolor{LightCyan}{rgb}{0.88,1,1}
\definecolor{LightGray}{rgb}{0.9,0.9,0.9}

\newcommand{\red}[1]{{#1}}
\begin{document}
\title{Learning from Noisy Labels with Decoupled Meta Label Purifier}



\author{Yuanpeng Tu$^{1}$\textsuperscript{*} \quad Boshen Zhang$^{2}$\textsuperscript{*} \quad Yuxi Li$^{2}$\textsuperscript{*} \quad Liang Liu$^{2}$ \quad Jian Li$^{2}$ \\ Yabiao Wang$^{2}$ \quad  Chengjie Wang$^{2}$ \quad Cai Rong Zhao$^{1}$\\
	$^{1}$Dept. of Electronic and Information Engineering, Tongji Univeristy, Shanghai \\ $^{2}$YouTu Lab, Tencent, Shanghai \\
	{\tt\small \{2030809, zhaocairong\}@tongji.edu.cn}\\
	{\tt\small \{boshenzhang, yukiyxli, leoneliu, swordli, caseywang, jasoncjwang\}@tencent.com}}

\maketitle

\footnote{$^{*}$ Yuanpeng Tu, Boshen Zhang, Yuxi Li contribute equally to this work.\\}

%

\newcommand{\fix}{\marginpar{FIX}}
\newcommand{\new}{\marginpar{NEW}}


\begin{abstract}
Training deep neural networks~(DNN) with noisy labels is challenging since DNN can easily memorize inaccurate labels, leading to poor generalization ability. Recently, the meta-learning based label correction strategy is widely adopted to tackle this problem via identifying and correcting potential noisy labels with the help of a small set of clean validation data. Although \lyx{training with purified labels can effectively improve performance}, solving the meta-learning problem inevitably involves a nested loop of bi-level optimization between model weights and hyper-parameters~(i.e., label distribution). \lyx{As compromise, previous methods resort to a coupled learning process with alternating update.} In this paper, we empirically find such simultaneous optimization over both model weights and label distribution can not achieve an optimal routine, consequently limiting the representation ability of backbone and accuracy of corrected labels. From this observation, a novel multi-stage label purifier named DMLP is proposed. DMLP decouples the label correction process into label-free \lyx{representation learning} and a \lyx{simple meta label purifier}, 
In this way, DMLP can focus on extracting discriminative feature and label correction in \lyx{two distinctive} stages. DMLP is a plug-and-play label purifier, \lyx{the purified labels} can be \lyx{directly reused in naive end-to-end network retraining or other} robust learning methods, where state-of-the-art results are obtained on several synthetic and real-world noisy datasets, especially under high noise levels.
\end{abstract}

\section{Introduction}

\begin{figure}[!t]
\centering
\includegraphics[width=0.4\textwidth]{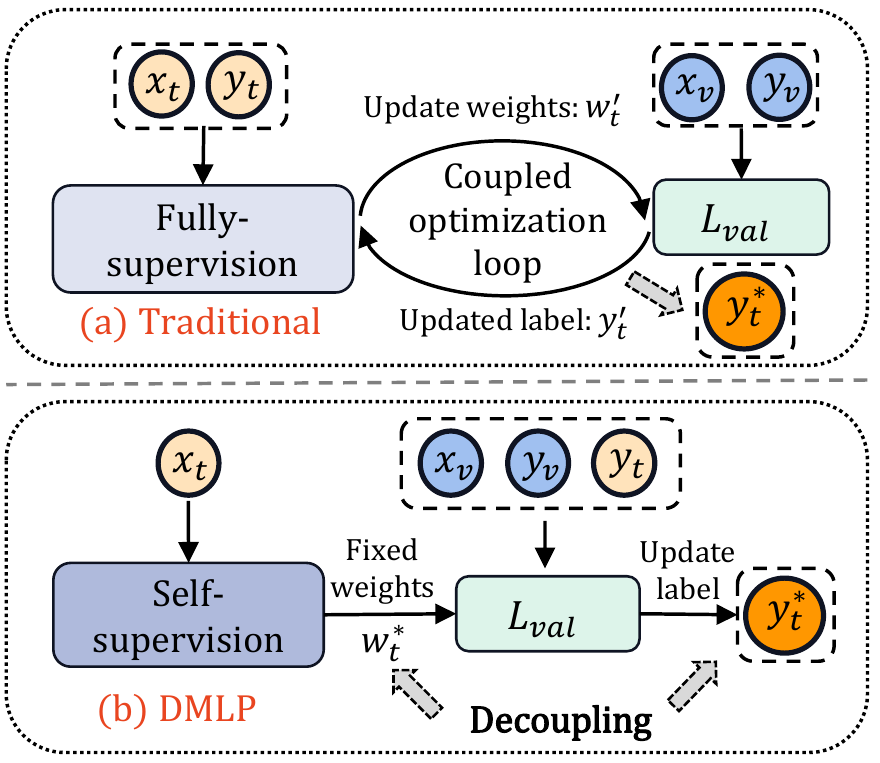}
\vspace{-0.8em}
\caption{ 
\leone{(a) \lyx{Traditional coupled alternating update to solve meta label purification problem}, and (b) the proposed DMLP method that decouples the \zbs{label purification} process into representation learning and a simple non-nested meta label purifier.}}
\vspace{-0.4em}
\label{fig:concept}

\end{figure}
Deep learning has achieved significant progress on various recognition tasks. The key to its success is the availability of large-scale datasets with reliable annotations. Collecting such datasets, however, is time-consuming and expensive. Easy ways to obtain labeled data, such as web crawling~\cite{Clothing1M}, inevitably yield samples with noisy labels, which is not appropriate to be directly utilized to train DNN since these complex models {are vulnerable to} memorize noisy labels~\cite{memorizationEffect}.

Towards this problem, numerous Learning with Noisy Label (LNL) approaches were proposed. Classical LNL methods focus on identifying the noisy samples and reducing their {effect on parameter updates by} abandoning~\cite{Co-han2018co} or assigning smaller importance. 
However, when it comes to extremely {noisy and complex scenarios, such scheme struggles since there is no sufficient clean data to train a discriminative classifier.} 
{Therefore}, label correction approaches {are proposed to augment clean training samples by revising} noisy labels to underlying {correct} ones. Among them, meta-learning based approaches~\cite{Learning-to-Reweight,MLNT,AAAI-2021-meta} achieve state-of-the-art performance via resorting to a small clean validation set {and taking noisy labels as hyper-parameters}, which provides sound guidance {toward} underlying label distribution of clean samples. However, {such \lyx{meta purification inevitably involves} a nested bi-level optimization problem on both model weight and hyper-parameters (shown as Fig.~\ref{fig:concept} \leone{(a)}), which is computationally infeasible. As a compromise, \lyx{the alternating update between model weights and hyper-parameters is} adopted to optimize the objective~\cite{Learning-to-Reweight,MLNT,AAAI-2021-meta},  \lyx{resulting in a coupled solution for both representation learning and label purification.}}

\begin{figure*}[!t]
\centering
\includegraphics[width=0.95\textwidth]{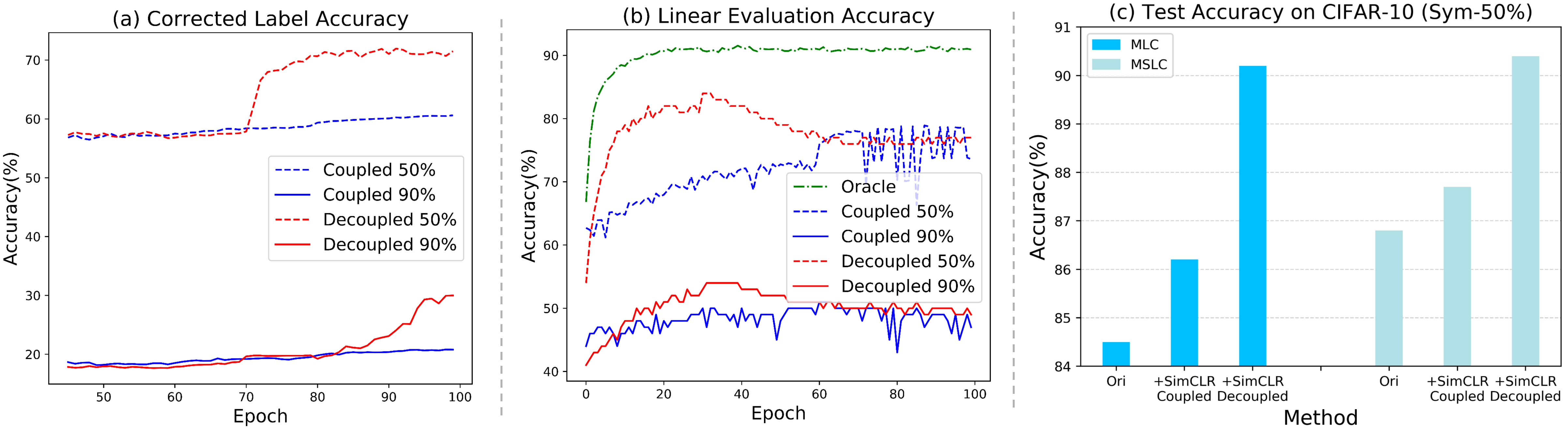}
\vspace{-1em}
\caption
    {
    \small
        \zbsN{The corrected label accuracy (a) and linear probe accuracy of representations (b) between coupled~\cite{zheng2021meta} and decoupled label correction schemes during training under 50\% and 90\% symmetric label noise on CIFAR-10. \zbs{(c) investigates the effect of representation
        learning on `Ori'-original coupled network training from scratch, `SimCLR-Coupled'-initialization with stronger self-supervised pretrained weights and `SimCLR-Decoupled'-further fix the pretrained backbone during label purification.}
        }  
      } 
  \label{fig:motivationNIPS}

\end{figure*}



{\textbf{Empirical observation.} Intuitively, alternate optimization over a large search space (model weight and hyper-parameters) may lead to sub-optimal solutions. To investigate how such approximation affects results in robust learning, we conduct empirical analysis on CIFAR-10~\cite{datasetcifar10} with recent label purification methods MLC~\cite{zheng2021meta} and MSLC~\cite{AAAI-2021-meta}, which consist of a deep model and a meta label correction network, and make observation as Fig.~\ref{fig:motivationNIPS}.}

\noindent $\bullet$ \textbf{Coupled optimization hinders quality of corrected labels.} We first compare the \emph{Coupled} meta corrector MLC with its extremely \emph{Decoupled} variant where the model weights are first optimized for $70$ epochs with noisy labels and get fixed, then labels are purified with the guidance of validation set. We adopt the accuracy of corrected label \lyx{to measure the performance of purification.} 
\zbsN{From Fig.~\ref{fig:motivationNIPS} (a), we can clearly observe that compared with \emph{Decoupled} counterpart, joint optimization yields inferior correction performance, and these miscorrection will reversely affect the representation learning in coupled optimization.}

\noindent $\bullet$ \textbf{Coupled optimization hinders representation ability.} We investigate the representation quality by evaluating the linear prob accuracy~\cite{2020MoCo-v2} of extracted feature in Fig.~\ref{fig:motivationNIPS} (b). We find the representation quality of \emph{Coupled} training is much worse at the beginning, which leads to slow and unstable representation learning in the later stage. To further investigate the effect on representation learning, we also resort to a well pretrained backbone with self-supervised learning~\cite{chen2020simple} as initialization, recent research~\cite{zheltonozhskii2022contrast} shows pretrained representation is substantially helpful for LNL framework. However, we find this conclusion does not strictly hold for coupled meta label correctors. As shown in Fig.~\ref{fig:motivationNIPS} (c), by comparing the classification accuracy from classifier of MLC/MSLC, we observe the pretrained model only brings marginal improvement if model weights is still coupled with hyper-parameters. In contrast, when the weight of backbone is fixed and decoupled from the label purification and classifier, the improvement becomes more significant.


{\textbf{Decoupled Meta Purification.}} \zbsN{From the observation above, } \lyx{we find the decoupling between model weights and hyperparameters of meta correctors is essential to label accuracy and final results. Therefore, }
in this paper, we {aim at detaching the meta label purification from representation learning and} \lyx{designing a simple meta label purifier which is more friendly to optimization of label distribution problem than existing complex meta networks~\cite{zheng2021meta, AAAI-2021-meta}.} Hence we propose a general multi-stage \lyx{label correction strategy}, named Decoupled Meta Label Purifier~(DMLP). 
{The core of DMLP is a meta-learning based label purifier, however,} \lyx{to avoid solving the bi-level optimization with a coupled solution}, DMLP decouples {this process into self-supervised} representation learning and {a linear meta-learner to fit underlying correct label distribution (illustrated as Fig.~\ref{fig:concept} \leone{(b)})}, thus simplifies the label \lyx{purification stage as} a single-level optimization problem. 
{The simple meta-learner is carefully designed with two mutually reinforcing correcting processes, named intrinsic primary correction (IPC) and extrinsic auxiliary correction (EAC) respectively. IPC plays the role of purifying labels in a global sense at a steady pace, while EAC targets at accelerating the purification process via looking ahead~(i.e., training with) the updated labels from IPC. The two processes can enhance the ability of each other and form a positive loop of label correction.}
{Our DMLP framework is flexible for application, \lyx{the purified labels} can either be directly applied for \lyx{naive end-to-end network retraining}, or \lyx{exploited} to boost the performance of existing LNL frameworks.} Extensive experiments conducted on mainstream benchmarks, including synthetic~(noisy versions of CIFAR) and real-world~(Clothing1M) datasets, demonstrate the superiority of {DMLP}. In a nutshell, the key contributions of this paper include:


$\bullet$ \lyx{We analyze the necessity of decoupled optimization for label correction in robust learning, based on which we propose }DMLP, a {flexible and} novel multi-stage \lyx{label purifier that solves bi-level meta-learning problem with a decoupled manner, which consists of} representation learning and {non-nested meta label purification; } 

$\bullet$ In DMLP, a novel non-nested meta label purifier equipped with two correctors, IPC and EAC is proposed. IPC is a global and steady corrector, while EAC accelerates the correction process via training with the updated labels from IPC. The two processes form a positive training loop {to learn more accurate label distribution};



$\bullet$ \lyx{Deep models trained with purified labels from DMLP} achieve state-of-the-art results on several synthetic and real-world noisy datasets across various types and levels of label noise, especially under high noise levels. Extensive ablation studies are provided to verify the effectiveness.

\section{Related Works}
The existing LNL approaches that are related to our work can be coarsely categorized into two groups: noisy sample detection and label correction.  

\textbf{Noisy sample detection} methods aim to identify and reduce the importance of suspicious false-labeled samples during training. The detected noisy samples are abandoned~\cite{Co-han2018co,aumranking2020}, assigned with smaller weights via a sample re-weight training scheme~\cite{importance-reweight-pami2015}, or used to formulate a semi-supervised learning problem by throwing away the labels while keeping the unlabeled data~\cite{li2020dividemix,zhang2020decoupling}. These methods show robustness under certain noise levels, but struggle when it comes to extremely noisy and complex scenarios since there is no sufficient clean data to train a classifier. 

\textbf{Label correction} approaches attempt to augment the training set by finding and correcting noisy labels to their underlying true ones. To do so, some works~\cite{patrini2017making} try to estimate the noise transition matrix. However, these methods usually assume that the noise type is class-dependent, which may be inappropriate for more complex noise settings, such as real-world noisy datasets~\cite{Clothing1M}.
Some other works resort to exploiting the prediction of the network, both soft~\cite{Reed2015Training,han2019deep,yi2019probabilistic,arazo2019unsupervised} and hard~\cite{tanaka2018joint,song2019selfie} label correction schemes are designed. However, the predictions of over-parameterized backbone network can be unreliable since it tends to fluctuate during training in the presence of false-labeled data~(\cite{zhang2020decoupling}). 
\zbsN{Another line of works utilize robust representations learned via unsupervised contrastive learning methods~\cite{zhang2020decoupling,li2021learning,ghosh2021contrastive,zheltonozhskii2022contrast} to eliminate the interference of noisy labels, which provides a reliable initialization of deep models.}
Recently, meta-learning based methods~\cite{Learning-to-Reweight,MLNT,AAAI-2021-meta,zheng2021meta} show great potential towards LNL problems with the help of a small clean validation set to provide sound guidance toward underlying label distribution of clean sample. However, these approaches \zbs{involves} a bi-level optimization problem on model weights and hyper-parameters, which is too computationally expensive to optimize. As a compromise, the one-step approximation is commonly adopted~\cite{AAAI-2021-meta,zheng2021meta} to convert the nested objective into a coupled update procedure between model weights and hyper-parameters, leading to sub-optimal performance.

Accordingly, DMLP belongs to the label correction group \zbs{via meta-learning strategy, but unlike previous meta-learning based methods, the learning process on model weights and labels are decoupled into individual stages within DMLP. Together with the proposed non-nested meta-label purifier, DMLP yields more accurate labels than coupled label correction methods~\cite{AAAI-2021-meta,zheng2021meta} and further set the new state-of-the-art on CIFAR and Clothing1M.}



\section{Method} \label{Method}
\subsection{Decoupled Solution to Meta Label Purification}
\begin{figure*}[!t]
\centering
\includegraphics[width=0.94\textwidth]{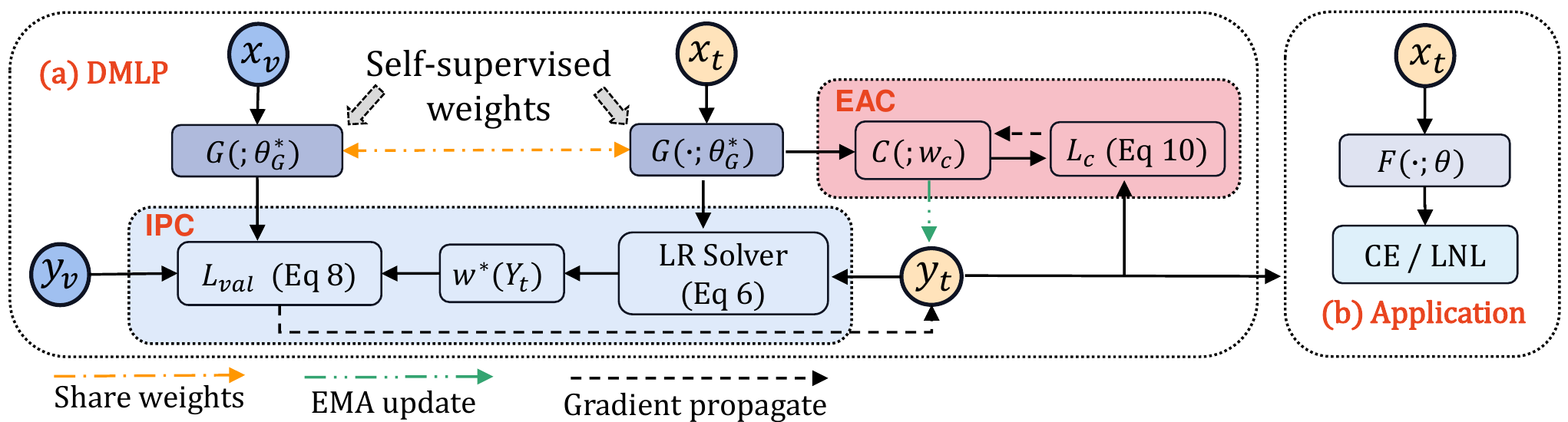}
\vspace{-3mm}
\caption{The overall framework of DMLP. (a) Decoupled meta label purification~(Sec. \ref{stage2}). (b) The purified labels can be applied in normal network retraining (CE) or other LNL loss~(Sec. \ref{stage3}).}
\vspace{-3.5mm}
\label{fig:framework}
\end{figure*}

For convenience, notations within DMLP are clarified first. The noisy training dataset is denoted as $D_t=\{(x_i,y_i)|1\leq i \leq N_t\}$, where $x_i \in \mathbb{R}^{H\times W\times 3}, y_i \in \{0,1\}^c$ are the image and corrupted label of the $i$-th instance, $c$ is the class number. Similarly, a small clean validation dataset is denoted as $D_v=\{(x_i,y_i)|1\leq i \leq N_v\}$. $N$ denotes the dataset size, {and $N_v \ll N_t $}. The subscripts $t$, $v$ represent the data is from training set or validation set. 

{Typical meta-learning based purification requires solving a bi-level optimization problem on both model weights and hyper-parameters with objective:}
\begin{equation}
\label{meta-objective}
\begin{gathered}
\min_{\theta_{\alpha}} E_{(x_v, y_v) \in D_v}\mathcal{L}_{val}(x_v, y_v; w^*(\theta_{\alpha})) \\
\textbf{s.t.}  \quad
w^{*}(\theta_{\alpha})=\arg \min _{w} E_{(x_t, y_t) \in D_t}\mathcal{L}_{train}(x_t, y_t; w, \theta_{\alpha}) \\
\end{gathered}
\end{equation} \\
where $w$ and $\theta_{\alpha}$ denote the model weights and the {meta hyper-parameters} respectively. {In the classical label purification pipeline, $\theta_{\alpha}$ is reparameterized as a function of noisy label distribution, i.e. $\theta_{\alpha}=g(y_t)$}. $\mathcal{L}_{val}$ and $\mathcal{L}_{train}$ are the loss function on different datasets. 
{Since both loss terms are \emph{not analytical} and involve complicated forward pass of DNN, solving the nested optimization objective is computationally expensive. Meanwhile, alternating one-step approximation~\cite{Learning-to-Reweight,AAAI-2021-meta} cannot guarantee that the optimization direction is optimal due to coupled update.}

{In contrast, in order to avoid coupled optimization over the large searching space of network parameters $w$, we reformulate the meta-learning objective in DMLP as:}
\begin{equation}
\label{eq:meta-objective-label}
\begin{aligned}
\min_{y_t} & E_{(x_v, y_v) \in D_v}\mathcal{L}_{val}(\mathbf{f}_v, y_v; w^*(y_t)) \\
\textbf{s.t.}  \quad
w^{*}(y_t) & =\arg \min _{w} E_{(x_t, y_t) \in D_t}\mathcal{L}_{train}(\mathbf{f}_t, y_t; w) \\
\mathbf{f}_t & = G(x_t;\theta^*_{G}), \quad \mathbf{f}_v = G(x_v;\theta^*_{G})
\end{aligned}
\end{equation} \\
where a pretrained feature extractor $G:\mathbb{R}^{H\times W\times 3}\rightarrow\mathbb{R}^d$ is designed to extract $d$-dimensional image representation $\mathbf{f}$. In order to obtain representation of high quality, these extracted features are utilized in a {contrastive} self-supervised learning {framework}~\cite{chen2020simple,2020MoCo-v2} to update the parameters $\theta_{G}$ as pretraining without noisy labels. Subsequently, the \zbsN{established} feature extractor $\mathbf{f}=G(x;\theta^*_G)$ can learn a noise-agnostic descriptor of image data, which is also highly separable in high dimensional feature space~\cite{zhang2020decoupling}. In this manner, we detach the representation learning from noisy label correction, while keeping strong separability of features.

{Further, since the feature is representative and separable, the loss term $\mathcal{L}_{train}$ and $\mathcal{L}_{val}$ can be formulated with simple  risk estimation functions (e.g. linear discrimination) instead of complex DNN forward pass, making it possible to solve the problem of Eq.~(\ref{eq:meta-objective-label}) in a non-nested manner with an \emph{analytical solution}, which will be introduced in Sec.~\ref{stage2}.}\\

\vspace{-.3em}
\subsection{Non-nested Meta Label Purifier} \label{stage2}
To solve the purification problem of Eq.~(\ref{eq:meta-objective-label}), we propose two mutually reinforced solutions to seek purified training labels as shown in Fig.~\ref{fig:framework}, the \textbf{i}ntrinsic \textbf{p}rimary \textbf{c}orrection (IPC) and \textbf{e}xtrinsic \textbf{a}uxiliary \textbf{c}orrection (EAC) processes.

\textbf{Intrinsic Primary Correction.} IPC \leone{aims} at performing global-wise label purification at a slow and steady pace.
Specifically, as shown in Fig.~\ref{fig:framework} (a), {the features and labels of a batch of $b$ training data are gathered into matrix $F_t \in \mathbb{R}^{b\times d}$ and $Y_t \in \mathbb{R}^{b \times c}$ as}
\begin{equation}
    F_t = [\mathbf{f}_{t,1},\mathbf{f}_{t,2},\cdots,\mathbf{f}_{t, b}]^T, \quad Y_t = [y_{t,1},y_{t,2},\cdots,y_{t,b}]^T.
\end{equation}

{Since the feature descriptors are representative, we assume there exists a simple linear estimation transform $w\in\mathbb{R}^{d\times c}$, which accurately predict the categorical distribution with ridge regression:}
\begin{equation}
\min_{w}\mathcal{L}_{train}(F_t,Y_t;w) = \left\|\sigma(\alpha Y_t)- F_{t}w\right\|^{2} + \lambda \left\|w\right\|^2,
\label{eq:leastori}
\end{equation}
where $\sigma(\cdot)$ is the softmax function along the categorical dimension to meet the normalization constraint, and $\alpha$ is a scaling factor. By {solving the linear regression problem of} Eq.~(\ref{eq:leastori}) through least square method, we can obtain its closed-form solution $w^{*}(Y_t)$ on the training {batch} $(F_t, Y_t)$ {and derive its optimal prediction via linear regression on samples $\mathbf{f}_{v,i}$} from validation set $D_{v}$:
\begin{align}\label{eq:predict}
{y^{\prime}}_{v,i}(Y_t) & = w^{*}(Y_t)^T\mathbf{f}_{v,i}\nonumber\\
                        & = \sigma(\alpha Y_t)^T{F_t}\left({F_t}^{T} {F_t} + \lambda I\right)^{-1}\mathbf{f}_{v,i}.
\end{align}
Intuitively, the discrepancy between the predicted results $y'_{v,i}$ and the ground truth labels $y_{v,i}$ of $D_v$ is due to the potential noise from $Y_t$, therefore we take the prediction discrepancy as objective for label purification:
\begin{equation}
{\mathcal{L}_{\rm val}}(Y_t)=\frac{1}{N_v}\sum_{i=1}^{N_v}{||{y^{\prime}}_{v,i}(Y_t)-y_{v,i}||^2+\mathcal{H}({y^{\prime}}_{v,i}(Y_t)),}
\label{eq:vallabels}
\end{equation}
{where $\mathcal{H}(\cdot)$ represents the entropy of input distribution as a regularization term \typ{to sharpen the predicted label distribution as ~\cite{yi2019probabilistic}}.} 
{ With Eq. (\ref{eq:predict}) and Eq.~(\ref{eq:vallabels}), the validation loss can be expressed analytically by training labels $Y_t$ in a batch, thus} the noisy labels can be corrected steadily {with correction rate $\eta_I$ by} the gradient back-propagated {from} Eq.~(\ref{eq:vallabels}) as
\begin{equation}
    {Y_t}^{p+1}:={Y_t}^{p}-\eta_I\nabla({\mathcal{L}_{\rm val}}(Y^p_t)).
    \label{eq:update_eaca}
\end{equation}

\begin{table*}[!t]
\small
    \centering
    \caption{
        Comparison with state-of-the-art methods  on CIFAR-10/100 datasets with symmetric noise. 
        ``CE'' is the standard ConvNet trained with Cross-Entropy loss in an end-to-end manner. \zbs{``Classifier'' means adopts the pre-trained SimCLR features to re-train a linear classifier. \red{``Val'' denotes using a small clean validation set.}}  DivideMix* denotes training DivideMix with the same validation set as additional data.}
    \vspace{-1em}
    
    \setlength\tabcolsep{10pt}
    \resizebox{0.9\textwidth}{!}{$
    \begin{tabular}{l c c |c|c|c|c||c|c|c|c}
        \toprule
        Dataset &      & & \multicolumn{4}{c||}{CIFAR-10}&\multicolumn{4}{c}{CIFAR-100}                    \\ 
        Method & \red{Val} & Noise ratio  & 20\%             &     50\%      &     80\%      &     90\%      &       20\%       &       50\%       &     80\%      &     90\%      \\ \midrule
        \multirow{2}{*}{Cross-Entropy (CE)}   &                                   & Best & 86.8             &     79.4      &     62.9      &     42.7      &       62.0       &       46.7       &     19.9      &     10.1      \\
                                                                  &       \red{\multirow{-2}{*}{\XSolidBrush}}      & Last & 82.7             &     57.9      &     26.1      &     16.8      &       61.8       &       37.3       &      8.8      &      3.5      \\ \midrule
        \multirow{2}{*}{Co-teaching$+$~\cite{yu2019does}}   &                   & Best & 89.5             &     85.7      &     67.4      &     47.9      &       65.6       &       51.8       &     27.9      &     13.7      \\
                                                      &             \red{\multirow{-2}{*}{\XSolidBrush}}             & Last & 88.2             &     84.1      &     45.5      &     30.1      &       64.1       &       45.3       &     15.5      &      8.8      \\ \midrule
        \multirow{2}{*}{PENCIL~\cite{yi2019probabilistic}}   &                & Best & 92.4             &     89.1      &     77.5      &     58.9      &       69.4       &       57.5       &     31.1      &     15.3      \\
                                                         &         \red{ \multirow{-2}{*}{\XSolidBrush}}              & Last & 92.0             &     88.7      &     76.5      &     58.2      &       68.1       &       56.4       &     20.7      &      8.8      \\ \midrule



        \multirow{2}{*}{REED~\cite{zhang2020decoupling}}        &               & Best & 95.8 & 95.6 & 94.3 & 93.6 & 76.7 & 73.0 & 66.9 & {59.6} \\
                                                                            & \red{\multirow{-2}{*}{\XSolidBrush}}   & Last & 95.7 & 95.4 & 94.1 & 93.5 & 76.5 & 72.2 & 66.5 & 59.4 \\  \midrule
                                                                               
                                                                               
                                                                               
        \multirow{2}{*}{Sel-CL+~\cite{li2022selective}}         &             & Best & 95.5 & 93.9 & 89.2& 81.9 & 76.5 & 72.4 & 59.6 & {48.8} \\
                                                                         &   \red{\multirow{-2}{*}{\XSolidBrush}}     & Last & 95.1 & 93.3 & 88.7 & 81.6 & 76.1 & 72.0 & 59.2 & 48.6 \\  \midrule
                                                                               
        \multirow{2}{*}{MOIT+~\cite{ortego2021multi}}                  &     & Best & 94.1 & 91.8 & 81.1 & 74.7 & 75.9 & 70.6 & 47.6 & {41.8} \\
                                                                       &      \red{\multirow{-2}{*}{\XSolidBrush}}   & Last & 93.8 & 91.3 & 80.6 & 74.0 & 75.2 & 70.1 & 46.9 & 41.2 \\  \midrule    
        \multirow{2}{*}{C2D-DivideMix~\cite{zheltonozhskii2022contrast}}                       & & Best & \textbf{96.3} & 95.2 & 94.4 & 93.5 & 78.6 & 76.4 & 67.7 & {58.7} \\
                                                                           &  \red{\multirow{-2}{*}{\XSolidBrush}}  & Last & \textbf{96.2} & 95.1 & 94.1 & 93.4 & 78.3 & 76.0 & 67.4 & 58.4 \\  \midrule
                                                                               
        \multirow{2}{*}{DivideMix~\cite{li2020dividemix}}  &        & Best & {96.1} &     {94.6}      &     {93.2}      &     76.0     &       {77.3}       &       {74.6}       &     60.2      &     31.5      \\
                                                                     &     \red{\multirow{-2}{*}{\XSolidBrush} }      & Last & {95.7} &     {94.4}      &     {92.9}      &     75.4     &       {76.9}       &       {74.2}       &     59.6      &     31.0      \\ \midrule
                                                      \midrule
   
        \multirow{2}{*}{Meta-Learning~\cite{MLNT}}      &                     & Best & 92.9             &     89.3      &     77.4      &     58.7      &       68.5       &       59.2       &     42.4      &     19.5      \\
                                                        &               \red{\multirow{-2}{*}{\CheckmarkBold} }         & Last & 92.0             &     88.8      &     76.1      &     58.3      &       67.7       &       58.0       &     40.1      &     14.3      \\ \midrule
        \multirow{2}{*}{MLC~\cite{zheng2021meta}} & & Best &   92.6        &   88.1        &  77.4         & 67.9        &   66.8           &   52.7           &  21.8        &  15.0         \\
                             &\red{\multirow{-2}{*}{\CheckmarkBold}}  & Last &  91.8         & 87.5      & 77.1         &  67.0       & 66.5            &  52.4         &  18.9         &  14.2        \\ \midrule
        \multirow{2}{*}{MSLC~\cite{AAAI-2021-meta}} & &Best & 93.4     & 89.9         & 69.8       &  56.1        &   72.5           &  65.4           &   24.3        & 16.7         \\
             & \red{\multirow{-2}{*}{\CheckmarkBold}}&Last & 93.3     & 89.4         & 68.8       &  55.2          & 72.0          &   64.9           &  20.5         & 14.6          \\ \midrule

        \multirow{2}{*}{DivideMix{*}~\cite{li2020dividemix}} & & Best &   96.1        &   94.9        &          93.6 &77.3         &     77.7         &   74.8           &60.7          &32.5           \\
                             &\red{\multirow{-2}{*}{\CheckmarkBold}}  & Last & 95.9          & 94.6      &  93.0        & 76.5        &77.1             &  74.3         &  60.5         &  32.2        \\

                    \midrule \midrule
        \multirow{2}{*}{DMLP-Naive}                           &  & Best & 94.7 & 94.2 & 93.5 &  92.8 &72.7 & 68.0 & 63.5 & 61.3  \\  
        &   \red{ \multirow{-2}{*}{\CheckmarkBold}}          & Last & 94.2 & 94.0 & 93.2 & 92.0 & 72.3 & 67.4 & 63.2 & 60.9 \\ \midrule                                                           
                                                                    
        \multirow{2}{*}{DMLP-DivideMix}         &                                       & Best & \textbf{96.3} & \textbf{95.8} & \textbf{94.5} & \textbf{94.3} & \textbf{79.9} & \textbf{76.8} & \textbf{68.6} & \textbf{65.8} \\
                                                 &               \red{\multirow{-2}{*}{\CheckmarkBold}}                & Last & \textbf{96.2} & \textbf{95.6} & \textbf{94.3} & \textbf{94.0} & \textbf{79.4} & \textbf{76.1} & \textbf{68.5} & \textbf{65.4}  \\ \bottomrule
    \end{tabular}$}
    \label{tab:cifar_sym}
\end{table*}

\textbf{Extrinsic Auxiliary Correction.} To accelerate the label correction process, {an external correction process} is {further} proposed. Specifically, an {accompanied linear} classifier $C(\cdot;w_c) : \mathbb{R}^d \rightarrow \mathbb{R}^c$ with learnable parameter $w_c$ is trained {along} with the updated labels from IPC:
\begin{equation}
   {\mathcal{L}_{\rm c}(w_c)}={\mathcal{L}_{\rm ce}}(C(\mathbf{f}_t;w_c),{{y}^{\prime}}_t)+{\mathcal{H}}(C(\mathbf{f}_t;w_c)),
\label{eq:classifier}
\end{equation}
where ${{y}^{\prime}}_t$ is the updated training label from IPC, {and $\mathcal{L}_{\rm ce}$ denotes the cross entropy loss function.}
Since the {accompanied linear} classifier is intrinsically robust to noisy labels {in $y^{\prime}_t$}~\cite{Reed2015Training}, it can quickly achieve high correction accuracy. {With this intuition}, the predicted results of the classifier $C(\mathbf{f}_t;w_c)$ are used {to correct labels periodically}, specifically, the update rule of $Y_t$ switches to momentum update every $T$ iterations:
\begin{equation}
    Y^{p+1}_t := (1-\eta_{E})Y_t^{p} + \eta_{E} C(F_t;w_c) \quad \textbf{if} \quad p = nT,
\label{eq:eac}
\end{equation}
{where $T$ and $\eta_E$ are the period and momentum for update, and $n$ is an arbitrary positive integer to denote the $n$-th time of EAC update. In a global sense, after $T$ iterations of training, EAC can quickly achieve locally optimal label estimation by mimic of gradually updated labels from IPC, which reversely} facilitates the label correction of IPC {by providing} cleaner training labels. Subsequently, 
IPC and EAC form a positive loop {and mutually improve the quality of label correction of each other}.

\subsection{Application of DMLP}\label{stage3}
{The DMLP is a flexible label purifier, the corrected training labels $y_t^*$ can be applied in different ways in robust learning scenarios, as shown in Fig.~\ref{fig:framework}.}

 \textbf{Naive classification network with DMLP.} {In a simple and direct manner, we can take} the purified labels to retrain a neural network with simple cross-entropy loss (CE), here we term this simple application as {DMLP-Naive}.

 \zbsN{\textbf{LNL framework boosted by DMLP.} Considering that there may still be a small number of noisy or miscorrected labels after purification, another effective way to apply DMLP is to take the purified labels as new training samples for the existing LNL framework. In this work, we extend some classic frameworks~\cite{li2020dividemix,2020ELR,2021CDR,Co-han2018co} with DMLP, and the boosted LNL methods are denoted {with prefix of ``DMLP-''} (\emph{e.g.} DMLP-DivideMix, DMLP-ELR+, etc.)}

\section{Experiments} \label{sec:expriment}
\subsection{Experimental Settings}

\begin{table}
    \footnotesize
    \caption{Evaluation results with asymmetric noise of different noisy ratio on CIFAR-10. \red{``Validation'' denotes the method exploits a small clean validation set.}}
    \vspace{-1em}
    \centering
    \begin{tabular}{p{3.2cm} |c |cc}
        \toprule
        \multirow{2}{*}{Method}  & & \multicolumn{2}{c}{Noisy ratio} \\ \cmidrule{3-4} 
        & \cellcolor{white}\multirow{-2}{*}{\red{Validation}} & 20\% & 40\% \\ \midrule
        Joint-Optim\cite{tanaka2018joint}       &   \red{{\XSolidBrush}}    & 92.8    & 91.7 \\ 
        
        PENCIL~\cite{yi2019probabilistic}       &    \red{{\XSolidBrush}}     & 92.4   & 91.2 \\ 
        M-correction~\cite{arazo2019unsupervised}   &  \red{{\XSolidBrush}}   &  -      & 86.3 \\ 
        Iterative-CV~\cite{chen2019understanding}    &  \red{{\XSolidBrush}}  &  -      & 88.0 \\ 
        DivideMix~\cite{li2020dividemix}  & \red{{\XSolidBrush}}  & 93.4    & 93.4 \\ 
        REED ~\cite{zhang2020decoupling}       &    \red{{\XSolidBrush}}      & {95.0}    & {92.3}    \\
        C2D-DivideMix ~\cite{zheltonozhskii2022contrast}     &       \red{{\XSolidBrush}}     &{93.8}    & {93.4}    \\
        Sel-CL+ ~\cite{li2022selective}       &      \red{{\XSolidBrush}}    & \textbf{95.2}    & {93.4}    \\
        GCE ~\cite{ghosh2021contrastive}      &      \red{{\XSolidBrush}}     & {87.3}    & {78.1}    \\
        RRL ~\cite{li2021learning}           &   \red{{\XSolidBrush}}   & -    & {92.4}    \\ 
        \midrule
        Zhang, et al.~\cite{zhang2020distilling}   &   \red{{\CheckmarkBold}}    & 92.7    & 90.2 \\
        Meta-Learning~\cite{MLNT}                &    \red{{\CheckmarkBold}}     &  -      & 88.6 \\ 
        MSLC ~\cite{AAAI-2021-meta} & \red{{\CheckmarkBold}}  &  94.4       & 91.6\\ 
        \midrule
        DMLP-Naive                      &          \red{{\CheckmarkBold}}            & 94.6    &93.9    \\
        DMLP-DivideMix                      &          \red{{\CheckmarkBold}}            & \textbf{95.2}    & \textbf{95.0}    \\
        \bottomrule
    \end{tabular}
    \label{tab:cifar10asy}
    \end{table}

\begin{table}
    \caption{Top-1 testing accuracy on Clothing-1M testset. \red{``Validation'' denotes using the validation provided by~\cite{Clothing1M}.}} 
    \vspace{-1em}
    \centering
    \footnotesize
    \begin{tabular}{l |c|c}
    \toprule        
        Method                          &  \red{Validation}           &  Top-1 Accuracy   \\ \midrule

        PENCIL~\cite{yi2019probabilistic}        &  \red{{\XSolidBrush}} &  73.49                 \\ 
        DivideMix~\cite{li2020dividemix}         &  \red{{\XSolidBrush}}  &  74.76                  \\ 
        RRL~\cite{li2021learning}  & \red{{\XSolidBrush}} & 74.90            \\ 
        GCE~\cite{ghosh2021contrastive}  & \red{{\XSolidBrush}} & 73.30          \\ 
        C2D-DivideMix~\cite{zheltonozhskii2022contrast}  & \red{{\XSolidBrush}} & 74.30             \\ 
        REED ~\cite{zhang2020decoupling}        &   \red{{\XSolidBrush}}  &  75.81    \\ \midrule
        Meta-Learning~\cite{MLNT}                & \red{{\CheckmarkBold}}   &  73.47                 \\ 
        Self-Learning~\cite{han2019deep}         &  \red{{\CheckmarkBold}}  &  76.44                  \\
        MLC ~\cite{zheng2021meta}& \red{{\CheckmarkBold}} &75.78                  \\ 
        MSLC ~\cite{AAAI-2021-meta}&\red{{\CheckmarkBold}} & 74.02                  \\ 
        Meta-Cleaner~\cite{zhang2019metacleaner} &  \red{{\CheckmarkBold}}  &  72.50                 \\ 
        Meta-Weight~\cite{shu2019meta} &  \red{{\CheckmarkBold}} & 73.72 \\
        FaMUS~\cite{xu2021faster} & \red{{\CheckmarkBold}} & 74.40 \\
        MSLG~\cite{algan2021meta} & \red{{\CheckmarkBold}} &76.02 \\
        \midrule
        DMLP-Naive                              &     \red{{\CheckmarkBold}}     &  77.77 \\
        DMLP-DivideMix                          &       \red{{\CheckmarkBold}}       &  \textbf{78.23}    \\     
        \bottomrule
    \end{tabular}
    \label{tab:clothing1M}
    \vspace{-2mm}
\end{table}

\textbf{CIFAR-10/100.} For the self-supervised pre-training stage, we adopt the popular SimCLR algorithm~\cite{chen2020simple} with ResNet as the backbone network. Classifiers in meta-learner are trained for 100 epochs with the Adam optimizer. For the final DivideMix algorithm, ResNet18 is adopted for a fair comparison. $\eta_{I}$ and $\eta_{E}$ are set as $0.01$ and $1.0$ respectively. The scaling factor $\alpha$ is set as $1.0$.
To ensure fair evaluation, we follow previous meta-learning based LNL methods~\cite{AAAI-2021-meta,shu2019meta} to randomly separate 1,000 images as the clean validation set for CIFAR-10/100~\cite{datasetcifar10}, {leaving the rest as training samples}. We strictly follow {the protocol in}~\cite{Co-han2018co} to generate label noise. Specifically, symmetric noise is generated by replacing labels with one of the other classes uniformly, while the labels in asymmetric noise are disturbed to their similar classes to simulate label noise in real-world scenarios. {Our experiments are conducted under different noisy rates}: $\pi \in \{20\%,50\%,80\%,90\%\}$ for symmetric and $\pi \in \{20\%,40\%\}$ for asymmetric noises.

\textbf{Clothing1M.} For the first stage on the Clothing1M~\cite{xiao2015learning}, the ResNet50 is trained with the official MoCo-v2~\cite{2020MoCo-v2} to fully leverage its advantages on large-scale datasets. Afterward, the {meta-learner is} trained for 50 epochs.{For DMLP-DivideMix}, ResNet50 is adopted and initialized with weights from previous stages and trained for 80 epochs.
Due to the space constraint, we put more detailed experimental settings and descriptions of each compared LNL method in our supplementary materials.

\subsection{Experimental Results}
 \textbf{Comparison with state-of-the-art methods.} We compare our method with multiple recent competitive methods on CIFAR-10/100 under various noisy settings (\zbs{detailed descriptions of these methods are provided in the supplementary materials}). Both test accuracy of the best and last epoch are reported. 
 As shown in Table ~\ref{tab:cifar_sym}, {the simple DMLP-Naive can already achieve competitive results to most methods, the superiority is more obvious in extremely noisy cases, further, the} DMLP-DivideMix achieves state-of-the-art performance across all the settings. 
\zbs{It is worth noting that directly utilizing the validation data to train DivideMix (i.e., DivideMix{*}) only brings marginal improvement, while when equipped with the purified labels by DMLP, significant improvements are obtained, indicating that DMLP is effective in terms of utilizing validation set towards LNL problem.}
{On the other hand}, though {there exist other} meta-learning methods {utilizing} validation set~\cite{MLNT,AAAI-2021-meta,zheng2021meta}, DMLP shows great advantages over them. {It is also noticeable that compared with the original DivideMix, the purified version of DMLP-DivideMix achieves better results by a large margin, indicating that the purified label from our approach is more friendly to boost LNL frameworks.} Table~\ref{tab:cifar10asy} shows comparison with the recent methods on asymmetric noisy CIFAR-10 dataset. DMLP-DivideMix outperforms REED by 0.2\% and 2.7\% {under different noisy ratios} and obtains greater improvements over the rest methods, {demonstrating} the ability of DMLP in handling harder {semantic-related} noise. Finally, DMLP-based methods suffer {less from increasing noisy ratio than other competitors,} 
indicating \emph{its robustness to variant noisy levels}.

\begin{table*}[!t]
\footnotesize
    \centering
    \caption{
    Comparison between the LNL methods and their DMLP applications with symmetric noise on CIFAR-10/100. Specifically, the 9-layer CNN is adopted as the backbone network of Co-teaching.}
    \vspace{-1em}
    \setlength\tabcolsep{10pt}
    \resizebox{0.8\textwidth}{!}{$
    \begin{tabular}{l l |c|c|c|c|c|c|c|c}
        \toprule
        Dataset      &      &  \multicolumn{4}{c}{CIFAR-10}       &  \multicolumn{4}{c}{CIFAR-100}            \\ 
        Method/Noise ratio &      & 20\% & 50\% &  80\%  & 90\%  & 20\% & 50\% &  80\%  & 90\%      \\ \midrule
        \multirow{2}{*}{Co-teaching~\cite{Co-han2018co}}  & Best & 82.6  & 73.0 & 24.0 & 14.6  & 50.5 & 38.2  & 11.8  & 4.9  \\  
                                                     & Last & 81.9  & 72.6 & 23.5 & 11.7 & 50.3 & 38.0  & 11.3 & 4.3\\ \midrule
        \multirow{2}{*}{DMLP-Co-teaching}            & Best & \textbf{85.8}  & \textbf{85.8} & \textbf{85.4} & \textbf{84.6}  &  \textbf{51.2} & \textbf{49.8}  & \textbf{48.1}  & \textbf{45.3}\\  
                                                     & Last & \textbf{85.6}  & \textbf{85.6} & \textbf{85.3} & \textbf{84.5}  & \textbf{51.0} & \textbf{49.3}  & \textbf{47.8} & \textbf{45.1} \\ \midrule\midrule      
        \multirow{2}{*}{CDR~\cite{2021CDR}}  & Best &90.4   &85.0    &47.2   & 12.3 & 63.3   & 39.5   &29.2   & 8.0   \\  
                                             & Last &82.7    &49.4   &16.6  &  10.1 & 62.9   & 39.5  &9.7   &  4.5 \\ \midrule
        \multirow{2}{*}{DMLP-CDR}            & Best & \textbf{91.4}   & \textbf{91.2}  & \textbf{91.2}   & \textbf{90.2} & \textbf{69.2}   & \textbf{64.8}  & \textbf{61.4}  & \textbf{58.5}  \\  
                                             & Last & \textbf{91.2}   & \textbf{90.8}  & \textbf{90.6}  & \textbf{89.3}   & \textbf{68.3}   & \textbf{64.3}  & \textbf{61.1}  & \textbf{57.9}   \\ \midrule\midrule
        \multirow{2}{*}{ELR+~\cite{2020ELR}}    & Best & 94.6 & 93.8 & 91.1 &  75.2 & 77.5 & 72.4 & 58.2 & 30.8 \\  
                                                & Last & 94.4 & 93.7 & 90.5 &  73.5 & 76.2 & 72.2  &56.8 &30.6  \\ \midrule
        \multirow{2}{*}{DMLP-ELR+}              & Best & \textbf{94.9} & \textbf{94.1} & \textbf{93.0} &  \textbf{92.5} & \textbf{77.8} & \textbf{73.6}  & \textbf{63.9} & \textbf{60.5} \\  
                                                & Last & \textbf{94.6} & \textbf{94.0} & \textbf{92.7} &  \textbf{92.1}  & \textbf{77.1} & \textbf{73.4}  & \textbf{63.6} & \textbf{60.5}   \\ 
                                             \bottomrule
    \end{tabular}$}
    \label{tab:cifar10_sym_compare}
\end{table*}

In addition to artificial noise, we also evaluate DMLP on the large-scale real-world noisy dataset Clothing1M. As shown in Table~\ref{tab:clothing1M}, {simple DMLP-Naive can} outperform all other methods by a large margin, {and DMLP-DivideMix further improves the accuracy by about $0.46\%$}. The results indicate that \emph{DMLP is more suitable for noise from real-world situations}. 

\begin{figure}[!t]
 \centering
\small
 \begin{minipage}{0.236\textwidth}
    \centering
    \includegraphics[width=\textwidth]{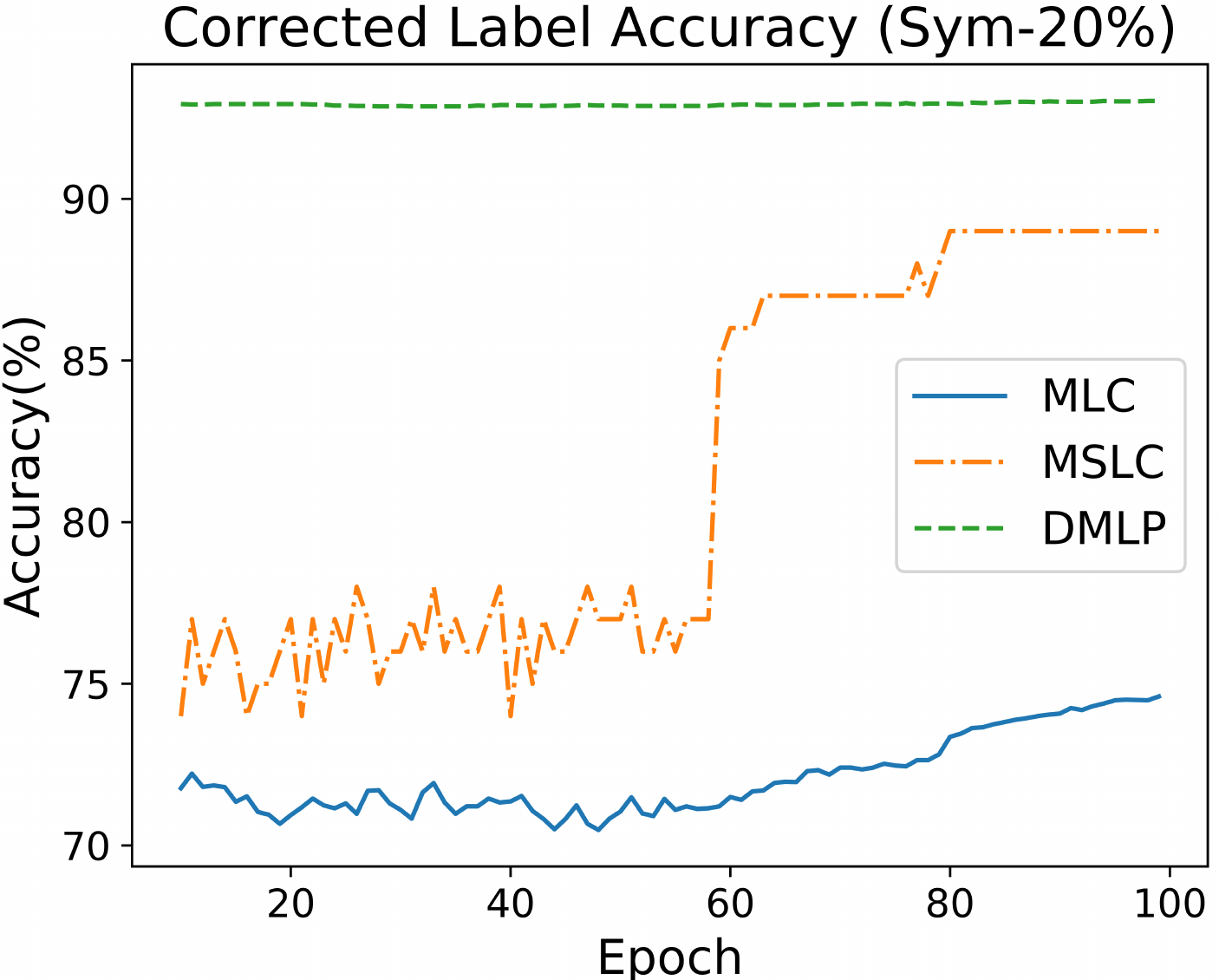}
\end{minipage}
 \begin{minipage}{0.236\textwidth}
    \centering
    \includegraphics[width=\textwidth]{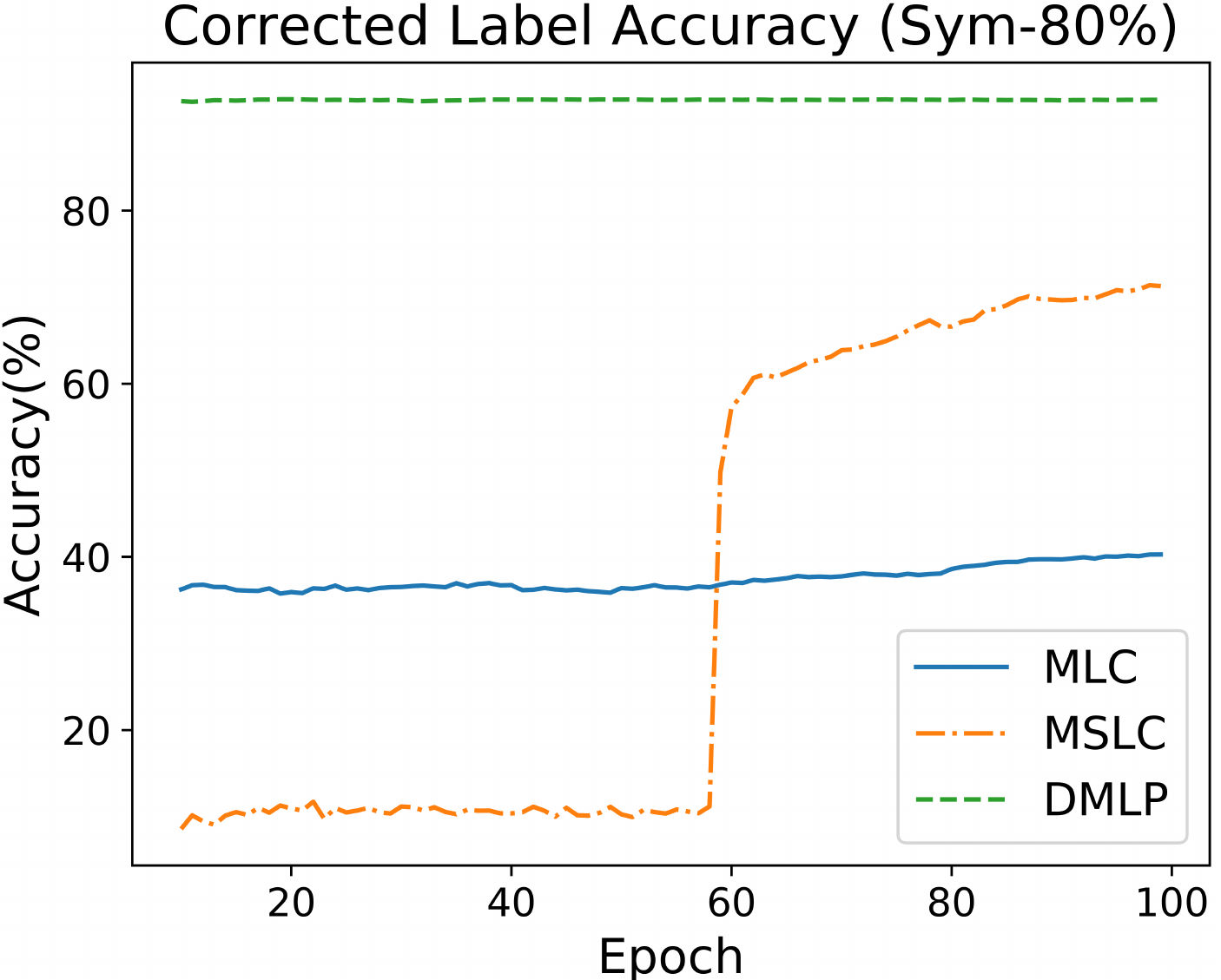}   
\end{minipage}
\vspace{-1em}
  \caption
    {
    \small
        Comparison of corrected label accuracy under symmetric-20\% (left), symmetric-80\% (right) noise settings on CIFAR-10. 
      }
  \label{fig:corrected_label}
  \vspace{-2mm}
 \end{figure}

 \textbf{Label correction accuracy.} Fig.~\ref{fig:corrected_label} \zbs{compares} the label accuracy {after correction in our meta-learner} \zbs{against coupled purifiers MLC and MSLC} on CIFAR-10. Specifically, the one-hot form of the corrected pseudo-labels is {compared} with the ground truth {for evaluation}. As Fig.~\ref{fig:corrected_label} shows, labels can be rapidly corrected to {accuracy} over 92\% in low noise cases. For severe label noise, DMLP can still improve label accuracy similar {to} low-noise \zbsN{settings}.
\zbs{\emph{The overall corrected label accuracy within DMLP is superior against other competitors across all noise settings} (More detailed experimental results can be found in the supplementary material). } 

 \textbf{Generality of DMLP.} 
\zbsN{ To validate the generalization ability of DMLP, other than DivideMix, another 3 popular LNL methods, ELR+~\cite{2020ELR}, Co-teaching~\cite{Co-han2018co}, and CDR~\cite{2021CDR} are further adopted to work collaboratively with purified labels of DMLP. As show in  Table~\ref{tab:cifar10_sym_compare},}
all the applications of DMLP perform consistently better over their corresponding baselines, especially under high-level noise cases. 
It is worth noting that since CDR highly relies on the early stopping technique, it suffers from a severe memorization effect in the training process, leading to a discrepancy between best results and last performance. In contrast, when training CDR with our purified labels, this discrepancy almost disappears, demonstrating the labels output by DMLP have a better quality to suppress the memorization effect, thus alleviating the reliance on early stopping. Therefore, the results indicate that \emph{the purified labels of DMLP are friendly to boost LNL frameworks.} {More detailed experimental results can be found in the supplementary materials.}

 \begin{table}[!t]
\footnotesize
    \caption{
        Ablation study for the effectiveness of IPC and EAC in DMLP-Naive on CIFAR-10. } 
    \vspace{-1em}
    \centering 
    \resizebox{0.49\textwidth}{!}{$
    \begin{tabular}{c c c|cccc|c}
    \toprule
        \multicolumn{3}{l|}{Component}                    &\multicolumn{4} {c|}{CIFAR-10} & \multirow{2}{*}{\makecell[c]{Clothing1M}} \\ 
                                          IPC & EAC   &            & 20\% & 50\% & 80\% & 90\%          \\ \midrule
        \multirow{2}{*}{$\xmark$}   & \multirow{2}{*}{\checkmark}  & Best &  93.7    &  93.3   & 91.1   & 67.4   & 76.5 \\
                                    &                              & Last &  93.0    &  92.9   & 90.6      & 66.5   & 76.1 \\ \midrule
        \multirow{2}{*}{\checkmark} & \multirow{2}{*}{$\xmark$}    & Best &  87.8    &  85.7   & 79.9   &   76.0 & 76.8 \\
                                    &                              & Last &  87.2     &  85.5   & 79.4   & 75.4   & 76.5 \\ \midrule
        \multirow{2}{*}{\checkmark} & \multirow{2}{*}{\checkmark}  & Best & \textbf{94.7} & \textbf{94.2} & \textbf{93.5} & \textbf{92.8} & \textbf{77.7}\\
                                    &                              & Last & \textbf{94.2} & \textbf{94.0} & \textbf{93.2} & \textbf{92.0} & \textbf{77.6}\\ 
    \bottomrule 
    \end{tabular} 
    $}
    \vspace{-3mm}
    \label{tab:ablation} 
\end{table} 
 
\subsection{Ablation Studies}
 \textbf{Analysis on different label correction process.} We explore the influence of
IPC and EAC on the performance of DMLP-Naive in Table~\ref{tab:ablation}, {when one of the process is excluded, the updated labels from the other are applied to retrain a new model}. {It is observed} that EAC performs well in low noise cases due to its intrinsic robustness, as the noise ratio increases, the performance drops rapidly. On the other hand, IPC is more robust to {high-level noise}, but there exists a large gap {compared with the full DMLP pipeline due} to its slow optimization process. In contrast, {when IPC and EAC work collaboratively, DMLP can achieve optimal results}. 



 

\textbf{\lyx{Comparison against other coupled purifiers with pretraining.}} To \lyx{solely} evaluate the influence \lyx{of decoupled purification}, we train two coupled meta label correction methods MLC and MSLC with \lyx{the same} self-supervised pretrained weights and apply their corrected labels to \lyx{naive training or} DivideMix for fair comparisons. As shown in Table.~\ref{tab:ablation_contrast}, though self-supervised weights can marginally boost \lyx{the performance of coupled label correctors}, there still exists a large gap between their performance and \lyx{DMLP}, especially for high noisy cases. Moreover, when further applying the corrected labels to mainstream LNL framework DivideMix, our method can also consistently outperform \lyx{the coupled counterparts} across all the noisy settings, demonstrating our corrected labels are of better quality. Therefore, these results verify that \emph{the superiority mainly attributes to decoupled label correction instead of self-supervised pretraining}.

\begin{table}
\small
\begin{minipage}{1.0\linewidth}
    \caption{
        Comparison with coupled meta label correction methods MLC~\cite{zheng2021meta} and MSLC~\cite{AAAI-2021-meta} on CIFAR-10. "*" denotes training with SimCLR pretrained ResNet-18.} 
    \vspace{-1em}
    \centering 
    \tabcolsep=1.5mm
    \begin{tabular}{c c |cccc}
    \toprule
        \multicolumn{2}{c|}{\multirow{2}{*}{Method}}&\multicolumn{4} {c}{Noisy ratio} \\ 
                      &                & 20\% & 50\% & 80\% & 90\% \\ \midrule
        \multirow{2}{*}{MLC*}     & Best &  91.8 & 86.2    & 77.6    &72.9  \\
                                  & Last &  91.6 & 85.9    & 77.5    &72.6  \\ \midrule
        \multirow{2}{*}{MSLC*}    & Best &  92.0 & 87.7    & 78.0    &67.8  \\
                                  & Last &  92.0 & 87.5    & 77.9    &67.3 \\ \midrule
        \multirow{2}{*}{DMLP-Naive*}& Best & \textbf{94.0} & \textbf{93.7} & \textbf{93.1} & \textbf{92.3} \\
                                    & Last & \textbf{93.9} & \textbf{93.4} & \textbf{92.9} & \textbf{91.9} \\  \midrule \midrule 
        \multirow{2}{*}{MLC*-DivideMix}& Best & 95.3  & 94.0   & 93.0   & 86.6 \\
                                       & Last & 95.0  & 93.6   & 92.7   & 86.5 \\ \midrule
        \multirow{2}{*}{MSLC*-DivideMix}& Best& 95.7  & 94.9   & 93.8 & 83.0  \\
                                        & Last& 95.5  & 94.8   & 93.1 & 82.8  \\ \midrule
        \multirow{2}{*}{DMLP*-DivideMix}& Best & \textbf{96.3} & \textbf{95.6} & \textbf{94.1} & \textbf{93.8} \\
                                       & Last & \textbf{96.0} & \textbf{95.2} & \textbf{94.0} & \textbf{93.6} \\    
    \bottomrule 
    \end{tabular} 
    \label{tab:ablation_contrast}

\end{minipage}
\end{table}

\begin{table}[!t]
\small
    \caption{
        Investigation of the validation set size $\tau$ on Clothing1M.} 
    \vspace{-1em}
    \centering 
    \setlength{\tabcolsep}{1mm}{
    \resizebox{0.43\textwidth}{!}{$
    \begin{tabular}{c |cccccc}
    \toprule
        $\tau$  & 10\% & 20\% & 30\% & 40\%    &50\% & 100\%      \\ \midrule
         Accuracy (\%) &   75.50 & 76.40 & 76.61 & 77.00 & 77.30 & \textbf{77.31} \\
    \bottomrule 
    \end{tabular} 
    $}}
    \label{tab:ablation_validnew} 
    \vspace{-4mm}
\end{table}


\textbf{Effect of different feature {representation for purification.}} The quality of features plays a crucial role in the label correction process of DMLP since {the distribution of learned features} is closely related to {the rationality behind the linear estimation assumption in high-dimensional space}. Therefore, we study the influence of different features on performance. Specifically, two types of features are investigated, including features from the ResNet-18/50 which load the self-supervised pre-trained weights. As the results in Table \ref{tab:ablation_feature} show, the features from the ResNet-18 lead to slightly poor performance, {while} it brings performance improvements when using features from self-supervised ResNet-50. {This observation indicates that although feature representation of higher quality benefits the purification results, DMLP is \emph{not very sensitive to the representation ability of input feature.}} 

 \textbf{{Effect of validation size}.} We examine how \zbs{the number of} validation set \zbs{affect performance}. Specifically, {validation} sizes \zbs{from 10\% to 100\% of the whole validation set} are evaluated on Clothing1M {for DMLP-Naive}. As shown in Table \ref{tab:ablation_validnew}, 
DMLP-Naive achieves similar performance regardless of the validation size $N_v$,
demonstrating DMLP is not sensitive to the number of images. \zbs{It is worth noting that even using only 10\% of the validation set (around 0.1\% of training data), DMLP-Naive still achieves high accuracy and outperforms most methods in Table.\ref{tab:clothing1M}, indicating that \emph{the effectiveness of DMLP is not heavily relied on validation size.}}

\textbf{Performance under extremely noisy setting.} {In an extremely noisy scenario where all labels in the training set are unreliable except the given clean validation set, the LNL problem is converted into a partially-labeled semi-supervised learning problem, therefore} we further compare DMLP-DivideMix with some state-of-the-art semi-supervised learning algorithms, including MeanTeacher~\cite{Meanteacher}, MixMatch~\cite{Mixmatch}, FixMatch~\cite{Fixmatch} and UDA~\cite{UDA}, under the 100\% symmetric noise case on CIFAR-10 and CIFAR-100. From the results in Table \ref{tab:ablation_semi}, DMLP-DivideMix performs optimally among all methods when using the validation set as labeled  samples for training, {indicating} {the proposed method is potential for {broader} applications}.

\begin{table}[]
\begin{minipage}{1.0\linewidth}
    \caption{
        Ablation study for adopting different features in DMLP-Naive on CIFAR-10, where "R18/50" denote "ResNet-18/50" and "M/S" represent "MoCo/SimCLR".} 
    \vspace{-1em}
    \centering 
    \tabcolsep=1.5mm
    \begin{tabular}{c c |cccc}
    \toprule
        \multicolumn{2}{c|}{\multirow{2}{*}{Feature Source}}                    &\multicolumn{4} {c}{Noisy ratio} \\ 
                      &                & 20\% & 50\% & 80\% & 90\% \\ \midrule
        \multirow{2}{*}{R18 (M)}     & Best &  93.8 & 93.3    & 92.2    &90.4  \\
                                     & Last &  93.7 & 92.7    & 92.1    &90.0  \\ \midrule
        \multirow{2}{*}{R18 (S)}     & Best &  94.0 & 93.7    & 93.1    &92.3  \\
                                     & Last &  93.9 & 93.4    & 92.9    &91.9   \\ \midrule
        \multirow{2}{*}{R50 (S)}     & Best & \textbf{94.7} & \textbf{94.2} & \textbf{93.5} & \textbf{92.8} \\
                                               & Last & \textbf{94.2} & \textbf{94.0} & \textbf{93.2} & \textbf{92.0} \\    
    \bottomrule 
    \end{tabular} 
    \vspace{-4mm}
    \label{tab:ablation_feature} 

\end{minipage}
\end{table}

\begin{table}[]
\begin{minipage}{1.0\linewidth}
    \vspace{1em}
    \caption{
       {Comparison between recent} semi-supervised methods {and DMLP-DivideMix} on CIFAR-10/100 with $100\%$ noisy ratio.} 
    \vspace{-1em}
    \centering 
    \setlength{\tabcolsep}{1mm}{
    \begin{tabular}{c|cc}
    \toprule
            Method & \makecell[c]{CIFAR-10}    & \makecell[c]{CIFAR-100} \\ \midrule
            MeanTeacher   & 83.0 & 31.0  \\
            MixMatch  & 87.9  & 57.7  \\
            FixMatch & 88.1 & 56.3  \\
            UDA    & 88.2 & 56.1  \\\midrule
            Ours & \textbf{91.7}  & \textbf{60.1}  \\
            
    \bottomrule 
    \end{tabular}}
    \vspace{-3mm}
    \label{tab:ablation_semi} 
\end{minipage}
\end{table}

\section{Conclusion}
In this paper, we propose a flexible and novel multi-stage robust learning approach {termed as} DMLP. The core of DMLP is a carefully-designed meta-learning based label purifier, which decouples {the complex bi-level optimization problem into} representation and label distribution learning, {thus helping} the meta-learner focus on correcting noisy labels in a faster and more precise manner even under extremely noisy scenarios. Further, DMLP can be applied either for naive retraining on noisy data or assistance of existing LNL methods to boost performance.
Extensive experiments conducted on several synthetic and real-world noisy datasets verify the superiority of the proposed method.

{\small
\bibliographystyle{ieee_fullname}
\bibliography{egbib}
}

\end{document}